\documentclass[letterpaper, 10 pt, conference]{ieeeconf}  % Comment this line out if you need a4paper

\IEEEoverridecommandlockouts                              % This command is only needed if 
\usepackage{graphicx}
\usepackage{epsfig} % for postscript graphics files
\usepackage{amsmath} % assumes amsm6666th package installed
\usepackage{mathtools}
\usepackage{bm}
\usepackage{amssymb}  % assumes amsmath package installed
\usepackage{subfigure}
\usepackage{algorithm}
\usepackage[noend]{algpseudocode}
\usepackage{multirow}
\usepackage{booktabs}
\usepackage{color}
\usepackage{url}

\newcommand{\forcef}{f}

\title{\LARGE \bf
    Manipulation Planning under Changing External Forces
}

\author{Lipeng Chen, Luis F. C. Figueredo, Mehmet Dogar% <-this % stops a space
\thanks{This project has received funding from the European Union's Horizon 2020 research and innovation programme under the Marie Sklodowska-Curie grants agreement
No. 746143 and 795714, and from the UK Engineering and Physical Sciences Research Council under grant EP/P019560/1.}% <-this % stops a space
\thanks{Authors are with School of Computing, University of Leeds, Leeds, UK,
        {\tt\small \{sclc, l.figueredo, m.r.dogar\}@leeds.ac.uk}}%
}

\begin{document}

\maketitle
\thispagestyle{empty}
\pagestyle{empty}

%%%%%%%%%%%%%%%%%%%%%%%%%%%%%%%%%%%%%%%%%%%%%%%%%%%%%%%%%%%%%%%%%%%%%%%%%%%%%%%%
\begin{abstract}
\textit{This work has been accepted and will appear in the 2018 IEEE/RSJ International Conference on Intelligent Robots and Systems (IROS 2018).}
	
In this work, we present a manipulation planning algorithm for a robot to keep an object
stable under changing external forces. We particularly focus on the case where
a human may be applying forceful operations, e.g. cutting or drilling, on an
object that the robot is holding. 
The planner produces an efficient plan by intelligently deciding when the robot
should change its grasp on the object as the human applies the forces. The planner also tries to choose
subsequent grasps such that they will minimize the number of regrasps that will be required in the
long-term.  Furthermore, as it switches from one grasp to the other, the planner solves the problem of bimanual regrasp
planning, where the object is \textbf{not} placed on a support surface, but
instead it is held by a single gripper until the second gripper moves to a new
position on the object. This requires the planner to also reason about the
stability of the object under gravity.  We provide an implementation on a
bimanual robot and present experiments to show the performance of our planner.
%
%In this paper we propose a virtual joint based method and experiments to check
%whether a grasp configuration can resist an external force. We also present
%planners and experiments for the multi-step manipulation planning problem under
%sequential external forces. We propose to represent the robot's grasp of a
%certain object with three prismatic virtual joints and three revolute virtual
%joints at each grip point. We perform experiments to measure the limits of
%these virtual joints and develop a grasp feasibility checking method based on
%the virtual joint limits and arm joint limits. The method can check whether a
%grasp configuration can resist a external force applied on a grasp object. We
%experimentally verify that the prediction of our virtual joint based grasp
%feasibility checking method is intentionally included within the data set of
%real experiments. We build a manipulation graph by sampling in the grasp space
%and checking the feasibility of each sample for the sequential forces. The
%graph is then used to solve the manipulation planning problem under a sequence
%of external forces by a simple graph search algorithm. Based on whether the
%planner can know the force sequences, we develop two planners to minimize the
%number of regrasps and  improving planning efficiency. We perform experiments
%to verify the effectiveness of the planners on drilling and cutting involved
%tasks using the Baxter robot.
%
\end{abstract}

%%%%%%%%%%%%%%%%%%%%%%%%%%%%%%%%%%%%%%%%%%%%%%%%%%%%%%%%%%%%%%%%%%%%%%%%%%%%%%%%
\section{Introduction}
\label{sec:intro} 

\begin{figure}[]
	\begin{center}
		\mbox{
			
			\subfigure[Robot Grasp 1]{{
					\label{fig:subfig:circlegrasp1}  
					\includegraphics[width=1.6 in, angle=-0]{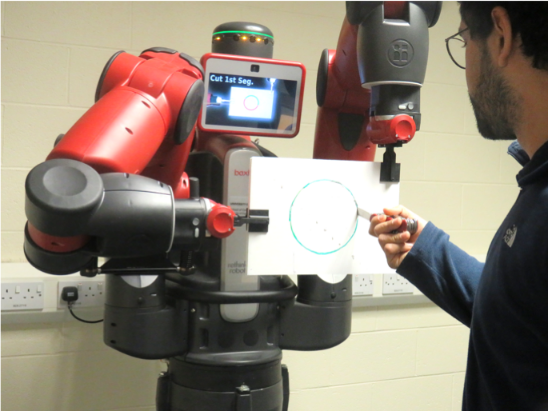}}}
			
			\subfigure[Intermediate config. for regrasp]{{
					\label{fig:subfig:circleinter}  
					\includegraphics[width=1.6 in, angle=-0]{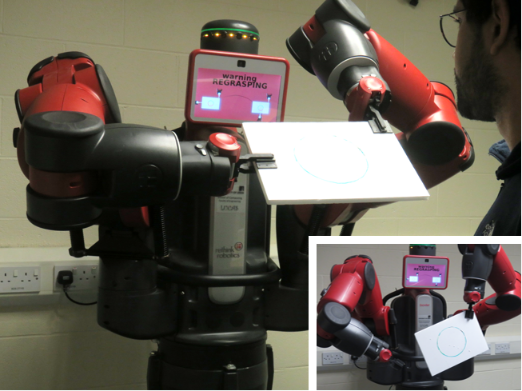}}}
		}
		
		\mbox{
			
			\subfigure[Robot Grasp 2]{{
					\label{fig:subfig:circlegrasp2}  
					\includegraphics[width=1.6 in, angle=-0]{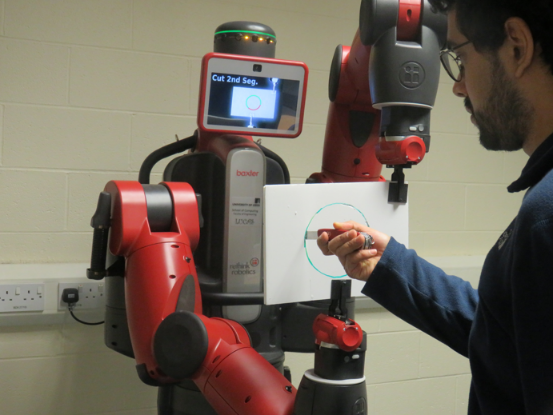}}}
				
			\subfigure[Robot Grasp 3]{{
					\label{fig:subfig:circlegrasp3}  
					\includegraphics[width=1.6 in, angle=-0]{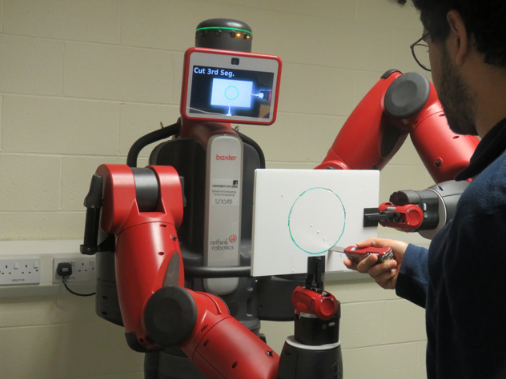}}}
			}

		\caption{Cutting a circular piece out of a board.}
		\label{fig:circlecut}
	\end{center}
\end{figure}

We are interested in the problem of a robot manipulating an object that is
under the application of changing external forces. Take the example in
Fig.~\ref{fig:circlecut}, where a human is cutting a circular piece out of a
board. During the cutting operation, the human exerts forces on the board that
change position, direction, and magnitude. To keep the object stable against
these forces, the robot changes its grasp on the object multiple times. 
In this paper we propose a planner that enables a robot to keep an object
stable under changing external forces like this.  

There are two key problems our planner solves. 

\begin{figure}[]
	\begin{center}
		\mbox{
			
			\subfigure[Object slides between fingers]{{
					\label{fig:failure_slip}
					\includegraphics[width=1.6 in, angle=-0]{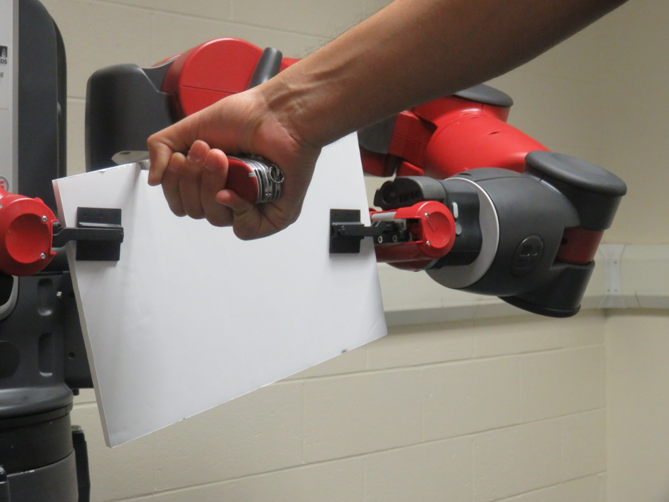}}}
			
			\subfigure[Object bends due to large torque]{{
					\label{fig:failure_bend}
					\includegraphics[width=1.6 in, angle=-0]{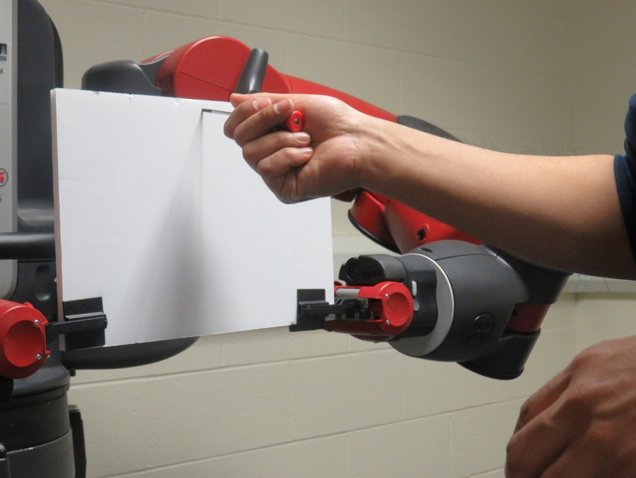}}}
		}
		
		\caption{Failure during cutting (a) and drilling (b).}
		\label{fig:failures}
	\end{center}
\end{figure}

First, our planner produces an efficient plan by minimizing the number of times
the robot needs to regrasp the object. For example in Fig.~\ref{fig:circlecut}, the robot 
changes its grippers' position only 2 times (counting each gripper separately)
during the whole operation.
This requires the planner to decide \textbf{when} to
regrasp during the course of the interaction.
It also requires the planner to choose grasps
intelligently.  A bad grasp may result in failure; for example the object may
slip through the fingers  during a cutting action
(Fig.~\ref{fig:failure_slip}), or it may bend away from the desired pose due to
large torques around the gripper during a drilling action (Fig.~\ref{fig:failure_bend}).

Second, our planner plans each regrasp. A regrasp requires the robot to release
its grippers off the object and then to grasp the object at different points.
However, when the robot releases a gripper, the object may become unstable
under external forces. Even if we assume the human in 
Fig.~\ref{fig:subfig:circlegrasp1} stops applying forces during regrasps, the object
can still become unstable due to gravity.  For example, to regrasp the object from the configuration in Fig.~\ref{fig:subfig:circlegrasp1} to the one in Fig.~\ref{fig:subfig:circlegrasp2},
if the robot simply releases its right gripper, a heavy object may slip within the remaining gripper as shown in the small figure at the right bottom of Fig.~\ref{fig:subfig:circleinter}. Therefore, the robot may need to
change the position of the object before releasing one of its grippers.
Fig.~\ref{fig:subfig:circleinter} shows such an intermediate pose, where the object is
stable even when the right gripper releases it.

In a typical \textit{multi-step manipulation planning} problem
\cite{simeon2004manipulation}, a robot moves an object through geometric
obstacles where the robot ungrasps and regrasps the object
multiple times. The need to regrasp objects was
recognized even in the earliest manipulation systems
\cite{lozano1987handey,tournassoud1987regrasping}. 
%Later, Sim{\'e}on et al.
% presented a planner based on randomized sampling
%that solves the problem using an alternating sequence of
%\textit{transfer} and \textit{transit} actions. 
More recently, planners have
been proposed to solve the regrasp planning problem in the case of multiple
manipulators for assembly-like tasks
\cite{wan2016developing,wan2016integrated,dogar2015multirobot,lertkultanon2017certified}.

We build on and extend this literature in three novel ways. 

First, in addition to
the kinematic and geometric (e.g. collision) constraints, we also consider
stability constraints due to the changing external forces acting on the
manipulated object. 
Multi-step manipulation planners need to go beyond geometric
constraints.
%and should be able to re-grasp objects also due to variable force constraints. 
In our task, for example, the robot is not
required to move the object to any goal position but is simply required to keep
the object stable. Still, due to the sequence of external forces acting on the object,
the robot needs to plan regrasps and the corresponding motions, possibly moving the object as a result. In this paper
we present such a manipulation planner.
Similar to Bretl \cite{bretl2006motion} we formulate the problem as first
identifying the stable intersections between different grasp manifolds and
then connecting these intersections.

Second, we solve the problem of regrasp planning ``in-the-air'' using two
manipulators. Existing work in regrasp planning focuses on placing an object
on a support surface and then regrasping it with a new gripper pose \cite{wan2016integrated,dogar2015multirobot,lertkultanon2017certified}. In our
task, the robot performs the regrasp without placing the object on a surface.
Instead, it goes through a sequence of unimanual and bimanual grasps to reach
different grasps.  This, however, requires our planner to also evaluate the
stability of the object against gravity, particularly during unimanual grasps.

Third, we are interested in addressing multi-step manipulation planning in a
\textit{human-robot interaction} setting.  Therefore we strive to minimize the
number of different grasps required to hold the object stable against external
forces.  We also have constraints in terms of where we position the object in
space to make it possible for the human to apply the forces.  Existing work in
forceful human-robot collaboration mostly focuses on the control problem
\cite{kosuge1997control,rozo2016learning,abi2017learning}, solving for the
necessary stiffness of manipulator joints as an external force is applied, and assumes the object to be already grasped at pre-specified points by the
robot.  We approach the problem from the manipulation
planning point of view and instead address the decision of what grasps
to use and when/how to switch between them.  Other work in planning for
human-robot collaboration exists \cite{luo2017unsupervised,maeda2017probabilistic,strabala2013towards} which focus on  
handing-over an object to a human, or
avoiding colliding a human working in the same workspace.  To
the best of our knowledge our work is the first one to take a planning approach
to the human-robot collaboration problem where the human applies multiple changing
forces on an object grasped by the robot.

\section{Problem definition} 
\label{sec:problem}

\begin{figure}[tbp]
	\begin{center}
		\mbox{
			
			\subfigure{{
					\label{subfig:forces}
					\includegraphics[width=2.2 in, angle=-0]{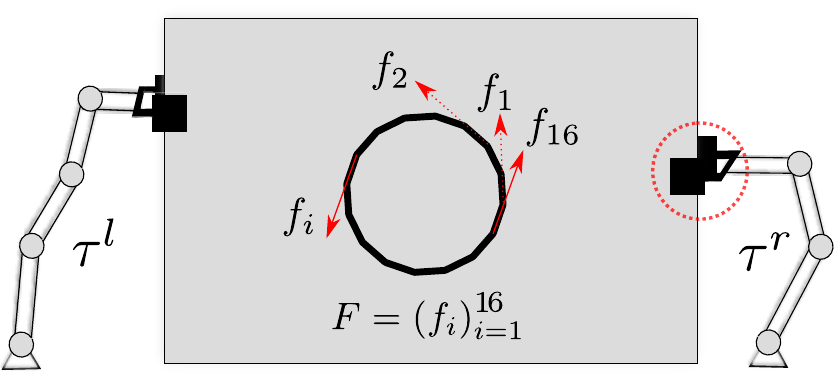}}}
			\subfigure{{
					\label{subfig:graph}
					\includegraphics[width=0.9 in, angle=-0]{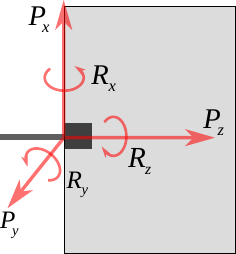}}}		
			
		}	
        \caption{Left: The task is represented as a sequence of forces $F=(\forcef_i)_{i=1}^{m}$. Right: We estimate the force/torque limits of a grip on the object along the three main axes.}
		\label{fig:forces}
	\end{center}
	\vspace{-5mm}
\end{figure}

In this paper, we are interested in scenarios where forces are exerted on the object grasped by
the robot. We use $\forcef$ to refer to a force vector, defined in the
object's coordinate frame.  Then, we represent a forceful task to be executed
on the object as a sequence of forces $F=(\forcef_i)_{i=1}^{m}$. For example, 
in Fig.~\ref{fig:forces}-Left,
a sequence of sixteen force vectors tangential to the circle represent the circular cutting task in Fig.~\ref{fig:forces}.

%Consider a 3D workspace with a robot and an object. 
Here, we assume the robot has two manipulators for clarity of explanation and
because the robot we use in our experiments has two arms. However, our formulation
can easily be extended to more manipulators.  We assume each manipulator is
equipped with a gripper.  
Let $CS^l$, $CS^r$ be the configuration space of the left and right
manipulator, and $SE(3)$ be the configuration space of the
object. The composite configuration space $CS$ is their Cartesian product ${CS
= CS^l \times CS^r \times SE(3)}$. Each composite configuration $q$ in $CS$ can
then be written as $q=(q^l,q^r,x)$, where $q^l \in CS^l$, $q^r \in CS^r$, and
$x \in SE(3)$. 

We also define a \textit{grasp}, $g$, using the pose of the gripper(s) on the
object. A bimanual grasp specifies poses for both the left and right grippers.
A unimanual grasp specifies the pose of only one gripper.  Such gripper poses
can be generated using a grasp planner, e.g. Miller and Allen
\cite{miller2004graspit}.  For example, the parallel plate grippers of the
Baxter robot, which we use in this work, can grip any point on the edges of the board.

A configuration $q$ and a grasp $g$ are related via forward/inverse kinematics.
Furthermore, the configuration space $CS$ consists of a collection
of lower-dimensional manifolds, where each manifold corresponds to a particular
unimanual or bimanual grasp of the object. We use $M(g)$ to refer to the
manifold for grasp $g$. For our planner, changing the grasp on the object means
changing the manifold the system is in. 

%By stability, we mean that the object should not slip or bend between the fingers
%(Fig.~\ref{fig:failure_examples}) and also that the arm joints should be able
%to generate the necessary torques to resist the force. 

In this paper, the robot's task is to stably grasp the object during the application of forces. 
%We would like to find a sequence of configurations $Q=(q_i)_{i=1}^p$ that are stable against the forces $F$.
Given a single force $\forcef$, we can
check whether the system is stable at a configuration $q$, using formulations from the literature in
grasp stability and cooperative manipulation (We explain how we perform this check in
Sec.~\ref{sec:stabilitycheck}). However, to reduce the number of regrasps required, the robot can use one
configuration against multiple external forces in a row.
In this work, we say that a
configuration $q$ is \textbf{stable against} a sequence of 
forces $(\forcef_i)_{i=1}^{m}$ if, at $q$, the system is stable against
    all $\forcef_i$. Moreover, we say that a sequence of configurations
$Q=(q_i)_{i=1}^p$ is stable against a sequence of forces
$F=(\forcef_i)_{i=1}^{m}$, if the configurations in $Q$ cover all the forces in $F$ in order, i.e. if $q_1$ is stable against
$(\forcef_1,...,\forcef_j)$, and $q_2$ is stable against
$(\forcef_{j+1},...,\forcef_k)$,  and so on until $q_p$ is stable
against $(\forcef_{n+1},...,\forcef_m)$, where $1 \leq j < k \leq n < m$.
For example the three configurations shown in Fig.~\ref{fig:circlecut} are stable against the forces distributed along a circle as shown in Fig.~\ref{fig:forces}.
Notice that different configurations correspond to different grasps on the object.

Finding a small set of configurations $Q=(q_i)_{i=1}^p$ to resist the forces is
only part of the problem. The robot must also be able to move between these
configurations, using collision-free and stable trajectories.

Therefore, given a sequence of external forces $F=(\forcef_i)_{i=1}^{m}$
and a starting configuration of the system $q_0$, we define the problem of
\textbf{manipulation planning under changing external forces} as the generation
of a sequence of configurations $Q=(q_i)_{i=1}^p$ and a sequence of
trajectories $T=(t_i)_{i=1}^{p}$, such that $Q$ is stable against $F$ and each
trajectory $t_i$ moves the system from $q_{i-1}$ to $q_i$, is collision-free
and is stable against gravity.  A trajectory $t_i$ usually corresponds to a
\textit{re-grasping} task.

Furthermore, we are interested in a human-robot interaction scenario. 
To make this interaction fluent for the human, we have the goal of 
 minimizing the number of regrasps required in the manipulation plan.  
 
In the human-robot interaction setting, we also assume a fixed desired pose of
the object, $x \in SE(3)$, that is comfortable for the human as he/she applies
forces on the object.  Therefore, we have the constraint that the
configurations $Q$ in the manipulation plan must position the object at $x$.

Hence, a planning query for us is a triple $(F,q_0,x)$ where $F$ is the
sequence of external forces to be applied on the object, $q_0$ is the starting
configuration of the system, and $x$ is the desired pose of the object when the
forces $F$ are applied.

\subsection{Overview of approach}

\begin{figure}[t!]
	\centering
	\includegraphics[width=\linewidth]{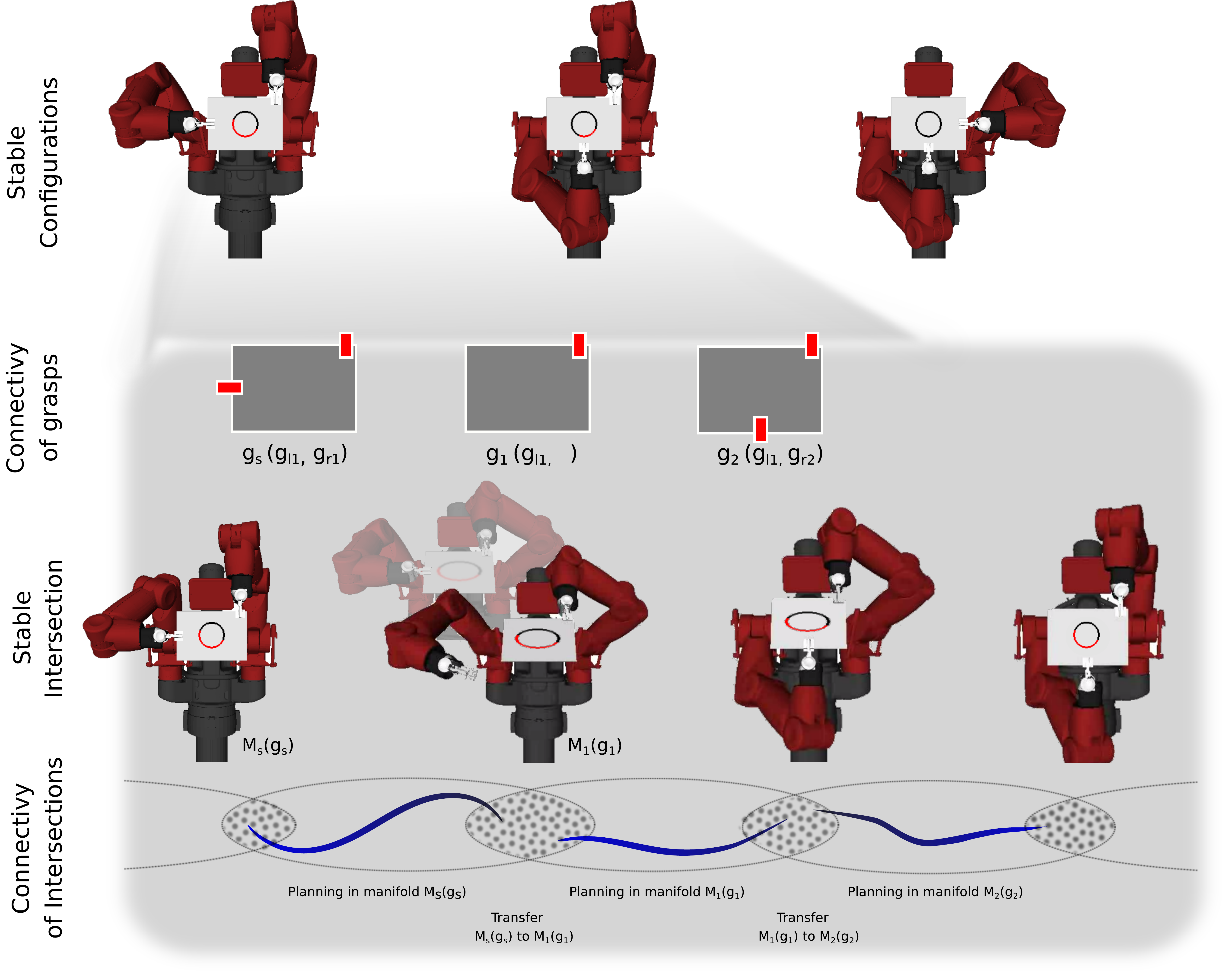}
	\caption{Overview of approach.}
	\label{fig:approach}
\end{figure}

\begin{figure*}[t!]
	%\begin{center}
		\mbox{
			\subfigure[Stable regions in $CS$]{{
					\label{subfig:cs}
					\includegraphics[width=0.45\linewidth]{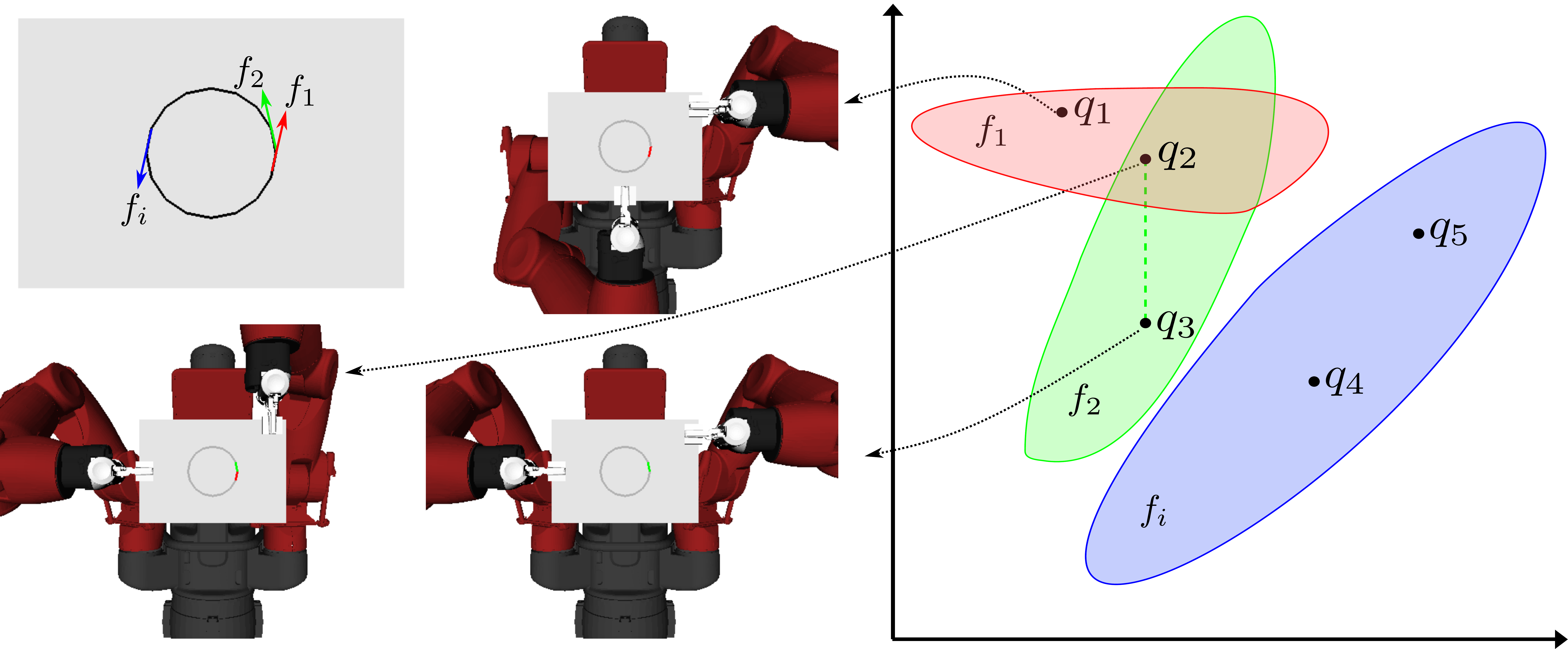}}}

                    \hspace{1mm}
			%\subfigure[Sampling]{{
			%		\label{subfig:samples}
			%		\includegraphics[width=1.6 in, angle=-0]{Manipulation/figs/sample.pdf}}}	
			
			\subfigure[Stable configs graph]{{
					\label{subfig:graph}
					\includegraphics[width=0.24\linewidth]{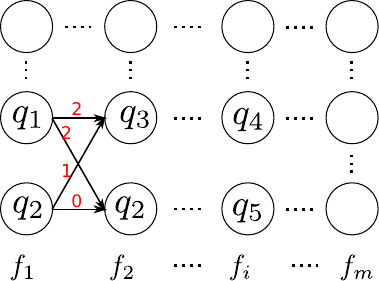}}}		
			
                    \hspace{3mm}
            \subfigure[Grasp connectivity graph]{{
					\label{subfig:regraspsolution}
					\includegraphics[width=0.24\linewidth]{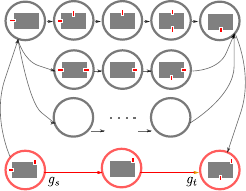}}}

		}	
		\caption{(a) and (b): Planning stable configurations. (c) Planning grasp connectivity.}
		\label{fig:graspspace}
	%\end{center}
	\vspace{-5mm}
\end{figure*}

Our problem is an instance of \textit{multi-modal planning}
\cite{hauser2010multi,bretl2006motion,lee2015hierarchical}, where each
different modality corresponds to a different bimanual or unimanual grasp.
In developing a planner, we follow a similar strategy of first identifying
intersection points between different modalities/manifolds, and then planning
motion paths to connect them.
We illustrate our overall planning approach in Fig.~\ref{fig:approach} in four
layers.  We present the details of each layer in Sec.~\ref{sec:approach}. Here
we present a brief overview and explain how these layers fit together:
\begin{itemize}
    \item \textit{Generating configurations stable against $F$.} Given $F$ and $x$, we first
        identify a candidate $Q=(q_i)_{i=1}^p$ which is stable against $F$, while 
        minimizing the number of regrasps. $Q$ also positions the object at
        $x$.  The three robot-object configurations shown in the top
        layer in Fig.~\ref{fig:approach} is an example output.  Given $Q$, the
        lower layers of the planner try to connect each subsequent
        configuration in $Q$.
    \item \textit{Connectivity of grasps.} Given two subsequent
        configurations generated in the top layer,
        $q_i$ and $q_{i+1}$, we identify a sequence of grasps $G=(g_j)_{j=1}^n$ on the object to move the grippers from their
        positions in $q_i$ to their positions in $q_{i+1}$. The second layer in
        Fig.~\ref{fig:approach} shows an example grasp sequence, 
        connecting the grasps in the first two configurations of the top layer.
        Note that there are many other
        possible contact sequences here, possibly going through other intermediate
        gripper contacts as shown in Fig.~\ref{subfig:regraspsolution}.
    \item \textit{Sampling stable intersections of grasp manifolds.} Given two subsequent
        grasps $g_i$ and $g_{i+1}$ from the sequence generated in the
        layer above, we sample a set of candidate configurations at which the
        transition from $g_i$ to $g_{i+1}$ can be performed
        stably.  The configuration in the middle on the third layer of the Fig.~\ref{fig:approach} is an
        example. At the shown configuration, both the unimanual grasp
        and the bimanual grasp can hold the object stable against gravity, and
        therefore this configuration is a good candidate to change between two
        grasp manifolds. 
    \item \textit{Connectivity of manifold intersections.} Given 
        a set of configurations at the intersections of sequences of grasp manifolds, this fourth layer
        performs collision-free and stability-constrained motion planning
        within the manifolds to connect the configurations.
        %the sequence
        %of configurations generated in the first layer and the third layer,
\end{itemize}

%If any of the layers fail in finding a solution, our planner backtracks to the
%layer above and searches an alternative candidate for the lower layer.  
The layered structure of our planner enables us to minimize the nuber of regrasps
at the top layer, but leaves the time-consuming motion planning to the final
layer, enabling fast planning time.

\section{Approach} 
\label{sec:approach} 

In this section we describe the details of our planner.

\subsection{Generating configurations stable against $F$}

Our planner takes input the sequence of forces $F$ and the desired object pose
$x$. It starts by generating a sequence of configurations $Q$ such that $Q$ is
stable against $F$, the configurations in $Q$ position the object at $x$, and
the number of regrasps between the configurations in $Q$ is minimized.

Given an external force $\forcef$, we can identify a set of configurations in
$CS$ which are stable against this force. In Fig.~\ref{subfig:cs}, the red, green and  blue regions illustrate such sets for forces $f_1$, $f_2$, and $f_i$ respectively. Note that there might be intersections between these stable
regions, and a configuration in the intersections is stable against multiple
forces; e.g. configuration $q_2$ in the figure is stable against
both $f_1$ and $f_2$.

Then our problem is to find a sequence $Q=(q_i)_{i=1}^p$ such that the configurations visit
the regions for each $f_i \in F$ in order. Moreover, we are interested in
identifying a sequence $Q$, such that it will minimize the needs for regrasping.

To create such a sequence of configurations $Q$, we first sample a set of
candidate configurations in $CS$.
To sample configurations that are likely to be stable against a variety of
forces, we start by sampling grasps on the object.
Using such a sampled grasp $g$ and the desired
object position $x$, we solve the inverse-kinematics problem, which may have
many solutions, and sample a single configuration $q$. 

For each sample $q$, we identify the
forces in $F$ that $q$ is stable against (Details of the stability check is explained below
in Sec.~\ref{sec:stabilitycheck}).  We then build a directed weighted graph
using these configurations as shown in Fig.~\ref{subfig:graph}.  In the graph, the
nodes in the ${i^{th}}$ column are the sampled configurations that are stable
against force ${\forcef_i}$. Then we create a link from every node in
${i^{th}}$ column to every node in ${(i + 1)^{th}}$ column.  We associate each link
with a weight using the number of gripper moves required from one configuration
to the other. For example, the weight between the node $q_2$ in the first
column and the node $q_2$ in the second column is zero, since the two nodes are
the same configurations and no re-grasping is needed.  Similarly, if two
configurations differ only by one gripper location on the object, the weight
for the link between them is set as one.  Otherwise, the weight would be two.
Note that one can come up with other weighting schemes, e.g., one that takes
into account the distance between grasp points.

At this point, our problem in this layer can be formulated
as graph search. We want a path that starts from one node in the leftmost
column for ${f_1}$ and ends with a node in the rightmost column for ${f_m}$.
We can search the graph for an optimal path. We use
\textit{Dijkstra's algorithm}, which gives us the
sequence $Q$ with the least number of gripper moves based on the current
set of samples. We call this planner the \textit{min-regrasp} planner. 

Building the graph requires knowing the sequence of external
forces $F$ beforehand. If the forces are revealed one by one, then the graph
can be formed as the next force is specified, and it can be searched greedily.
We call this version the \textit{greedy planner}.

We provide the pseudo-code for this layer of our planner in Alg.~1 in the
procedure $PlanStableSequence$. On line 1, we generate the graph as described
above. One line 2, we search this graph (e.g., Dijkstra's) to generate $Q$. Then
we iterate over every subsequent pair of configurations in $Q$ (line 4), and
try to plan a regrasp between them, which is explained below. If the regrasp
planning fails between two configurations (line 6), we remove the failing link
from the graph in Fig.~\ref{subfig:graph} (line 7), and re-search the graph
to generate a new $Q$ (line 8).

\begin{algorithm}[h!]
    \small
%\ContinuedFloat
\caption{Manipulation planning under changing forces} 
\label{alg:alg1}

\hspace*{0.0in} {\bf} 
$PlanStableSequence\left(F,q_{0},x\right)$:
\begin{algorithmic}[1]
    \State $V,E \leftarrow$ Sample configs and build graph in Fig.~\ref{subfig:graph}
    \State $Q \leftarrow GraphSearch(V,E)$
    \State $Q \leftarrow $ Append $q_0$ to beginning of $Q$
    \For {each subsequent $q_{i-1}$ and $q_{i}$ in $Q=(q_i)_{i=1}^{p}$}
        \State $t_i \leftarrow PlanRegrasp(q_{i-1},q_{i})$
        \If {$PlanRegrasp$ failed}
            \State $V,E \leftarrow$ Remove failing edge from graph $V,E$ 
            \State Go to line 2
        \EndIf
	\EndFor
    \State \Return ($Q=(q_i)_{i=1}^{p}$,$T=(t_i)_{i=1}^{p}$)
\end{algorithmic}

\vspace{0.05in}
\hspace*{0.0in} {\bf} 
$PlanRegrasp\left(q_{s},q_{t}\right)$:
\begin{algorithmic}[1]
    \State $V,E \leftarrow$ Sample grasps and build graph in Fig.~\ref{subfig:regraspsolution}
    \State $G \leftarrow GraphSearch(V,E)$
    \State $t \leftarrow Connect(q_s,G=(g_i)_{i=1}^{n},q_t)$
    \If {$Connect$ failed}
        \If {maximum number of attempts reached}
            \State \Return failure
        \EndIf 
        \State $V,E \leftarrow$ Remove failing edge from graph $V,E$ 
        \State Go to line 2
    \Else
        \State \Return $t$
    \EndIf
\end{algorithmic}

\vspace{0.05in} \hspace*{0.0in} {\bf}  
$SampleIntersection\left(g,g'\right):$
\begin{algorithmic}[1]
\State One of $g$ and $g'$ must be bimanual. Assuming $g$.
\State $S\leftarrow \left\{\right\}$
\While{$S$ has less than $m$ samples} 
    \State $x \leftarrow$ Sample pose for object
    \State $q \leftarrow$ Solve IK with object at $x$ and grippers at $g$
    \If {$q$ is stable against gravity with both $g$ and $g'$}
      \State Add $q$ to $S$
    \EndIf
\EndWhile
\State \Return $S$
\end{algorithmic}

\vspace{0.05in}
\hspace*{0.0in} {\bf} 
$Connect\left(q_s,G=(g_1,g_2,...,g_n),q_t\right):$
\begin{algorithmic}[1]
\If {$n = 1$} 
    \State $t \leftarrow MotionPlan(q_s,q_t)$ using grasp $g_n$
    \If {$MotionPlan$ successful}
        \State \Return $t$
    \Else
        \State \Return failure
    \EndIf
\EndIf
\State $S \leftarrow SampleIntersection(g_1,g_2)$
\For {each $q$ in $S$}
    \State $t \leftarrow MotionPlan(q_s,q)$ using grasp $g_1$
    \If {$MotionPlan$ successful}
        \State \Return $t + Connect(q,G=(g_2,...,g_n),q_t)$
    \EndIf
\EndFor
\State \Return failure
\end{algorithmic}

\end{algorithm}

\subsubsection{Stability check}\label{sec:stabilitycheck}

Given an external force $\forcef$, a configuration of the robot-object system
$q$, and the gripper contacts on the object, we check the stability of the
system against $\forcef$.
Given an external force on an object grasped by two cooperating manipulators,
the \textit{cooperative manipulation} literature provides formulations to compute
possible torque distributions on the manipulators' joints.  
Particularly, Uchiyama et al. provide the symmetric formulation
\cite{uchiyama1988symmetric,uchiyama1992symmetric}, which
describes the kinematic and static
relationship between the force applied on the object and
its counterparts required at the manipulator joints to resist it.
This formulation, however, leaves the forces at the grip points 
unconstrained. In addition to the manipulator joint torque limits, we are also interested
in checking whether the grip forces, e.g. the frictional forces between
fingers, will be able to resist the external force.
This requires the computation of the \textit{grasp wrench space}
\cite{mishra1987existence}, which is the space of all external wrenches a grasp
on an object can stably resist. 
For the parallel plate grippers we use in this work, we approximate the grasp wrench space
with an axis-aligned box in the six-dimensional force-torque space, i.e.
as maximum force and torque limits along each of the three main axes around a
grip point as shown in Fig.~\ref{fig:forces}-Right, where $\left[{{P_x},P{}_y,{P_z},{R_x},{R_y},{R_z}} \right]$ are these estimated
limits.

Imposing this additional constraint onto the symmetric formulation of  Uchiyama
et al. \cite{uchiyama1988symmetric,uchiyama1992symmetric}, we have the problem:
\begin{equation}
\label{eq:virtual_cons}
\begin{array}{l}
{J^T}{f^{g}} =  \tau \\
W{f^g} =  - f\\
\left| \tau \right| \le {\tau _{\max }}\\
\left| {{f^g}} \right| \le {f^g_{{{\max }^{}}}}
\end{array}
\end{equation} 
where
\small
\begin{equation*}
\begin{split}
		&J=
        \begin{bmatrix}
            J^l & 0       \\
            0   & J^r       
        \end{bmatrix} 
        ,\,
        {f^g}
        = 
        \begin{bmatrix}
            {f^{{g^l}}}        \\
            {f^{{g^r}}}       
        \end{bmatrix} 
        ,\,
        \tau
        = 
        \begin{bmatrix}
            \tau^l        \\
            \tau^r        
        \end{bmatrix}
        ,
       \\ 
        &\tau_{\max}
        = 
        \begin{bmatrix}
            \tau_{max}^l        \\
            \tau_{max}^r        
        \end{bmatrix} 
        ,\,
        f^g_{\max}
        = 
        \begin{bmatrix}
            f^{gl}_{max}        \\
            f^{gr}_{max}       
        \end{bmatrix} 
    %\]
\end{split}
\end{equation*} 
\normalsize
and
\begin{itemize}
\item ${J^l}$ and $J^r$ are the Jacobians of the two manipulators at the configuration we are checking the stability;
\item $f^{{g^l}}$ and $f^{{g^r}}$ are the forces and torques at the grippers of the two manipulators;
\item ${\tau ^l}$ and ${\tau ^r}$ are the vectors of torques acting at the joints of two manipulators;
\item $W$ (sometimes termed the grasp matrix \cite{mishra1987existence,ferrari1992planning,borst2004grasp}) is a $\left( {6 \times 12} \right)$ matrix mapping the forces and torques at the grippers to a resultant force on the object;
\item $\forcef$ is the external force/torque vector on the object;
\item $\tau _{\max }^l$ and  $\tau _{\max }^r$ are the torque limits at the joints of the manipulators;
\item $f^{gl}_{\max }$ and $f^{gr}_{\max }$  are our estimates of the maximum force and torque limits along each of the three main axes of each gripper (i.e. our estimate of the grasp wrench space): $f^{gl}_{\max } = f^{gr}_{\max } = \left[{{P_x},P{}_y,{P_z},{R_x},{R_y},{R_z}} \right]$.
%\item ${J^T} = {\rm{Blockdiag}}(\mathop {{J^l}}\nolimits^T , \mathop {{J^r}}\nolimits^T )$, and ${J^l}$ and $J^r$ are the Jacobians of the two manipulators at the configuration we are checking the stability;
%\item ${f^g} = {[{f^{{g^l}}}, {f^{{g^r}}}]^T}$, and $f^{{g^l}}$ and $f^{{g^r}}$ are the forces and torques at the grippers of the two manipulators;
%\item $\tau  = {[{\tau ^l}, {\tau ^r}]^T}$, and ${\tau ^l}$ and ${\tau ^r}$ are the vectors of torques acting at the joints of two manipulators;
%\item $W$ (sometimes termed the grasp matrix \cite{mishra1987existence,ferrari1992planning,borst2004grasp}) is a $\left( {6 \times 12} \right)$ matrix mapping the forces and torques at the grippers to a resultant force on the object;
%\item $\forcef$ is the external force/torque vector on the object;
%\item ${\tau _{\max }} = {\left[ {\tau _{\max }^l, \tau _{\max }^r} \right]^T}$, and $\tau _{\max }^l$ and  $\tau _{\max }^r$ are the the torque limits at the joints of the manipulators;
%\item ${f^g_{\max }} = {\left[ {f^{gl}_{\max }, f^{gr}_{\max }} \right]^T}$, and $f^{gl}_{\max }$ and $f^{gr}_{\max }$  are our estimates of the maximum force and torque limits along each of the three main axes of each gripper (i.e. our estimate of the grasp wrench space): $f^{gl}_{\max } = f^{gr}_{\max } = \left[{{P_x},P{}_y,{P_z},{R_x},{R_y},{R_z}} \right]$.
%
\end{itemize}

Eq.~\ref{eq:virtual_cons} is a linear programming problem, and can be solved,
e.g. using the Simplex method, to see if there are any feasible
solutions of the torques at the joints $\tau$ and forces/torques at the grip
points ${f^g}$. If this fails, we consider the configuration unstable against
the external force.

%Uchiyama et. al provide the well-known symmetric formulation \cite{uchiyama1988symmetric,uchiyama1992symmetric}:
%\begin{equation}
%\label{eq:cooperative}
%\begin{array}{l}
%{J^T}{f^{\rm{g}}} =  \tau \\
%W{f^g} =  - f\\
%\left| \tau \right| \le {\tau _{\max }}
%\end{array}
%\end{equation} 

%Eq.~\ref{eq:cooperative} 
%mathematically describes the kinematic and static
%relationships between the generalized external force applied on the object and
%its counterparts required at the manipulator joints and grippers to resist it.
%It 
%is a linear programming problem, and can be solved for example using the
%Simplex method, to see if there are any feasible solutions of the forces and
%torques ${f^g}$ at the grip points and the corresponding torques $\tau$ at the
%manipulator joints to resist the external force.  
%%Since $W$ usually has a non-trivial null space, there may be
%%infinitely many solutions.

%where ${f_{\max }} = {\left[ {f_{\max }^1, f_{\max }^2} \right]^T}$  and each
%$f_{\max }^i$ is a $\left( {6 \times 1} \right)$ vector $f_{\max }^i = \left[
%    {{P_x},P{}_y,{P_z},{R_x},{R_y},{R_z}} \right]$ to represent the force and
%    torque limits of ${i^{th}}$ gripper. This linear programming problem,
%    again, can be solved using the Simplex method to see if there are any
%    feasible distribution of the joint torques \textit{and} gripper forces,
%    such that both joint torque limits and the grippers' limits are respected.

%%%%%%%%%%%%%%%%%%%%%%%%%%%%%%%%%%%%%%%%%%%%%%%%%%%

\subsection{Connectivity of grasps}
\label{sec:connectivityofgrasps} 

%\begin{figure}[bthp]
%	\begin{center}
%		\subfigure{{
%				\label{subfig:regraspsolution}
%				\includegraphics[width=3.1 in, angle=-0]{Regrasp/fig/regrasp.pdf}}}
%		%\subfigure[$g_{s}\rightarrow g_{t}.$]{{
%		%		\label{subfig:prob}
%		%		\includegraphics[width=3. in, angle=-0]{Regrasp/fig/path2}}}
%		
%		\caption{We generate and search a grasp graph to identify a sequence of grasps connecting $g_{s}$ to  $g_{t}$.}
%		\label{fig:regraspsolution}
%	\end{center}
%	\vspace{-5mm}
%\end{figure}

Given two subsequent configurations generated in the previous layer, $q_i$ and
$q_{i+1}$ in $Q$, and their corresponding grasps $g_s$ and $g_t$, we identify a
sequence of grasps $G=(g_j)_{j=1}^{n}$ on the object to move the grippers from
$g_s$ to $g_t$.
For example, take the first two configurations in the top row of
Fig.~\ref{fig:approach}. The robot must go through a number of intermediate
grasps to move between the two grasps on the object (These intermediate grasps serve as alternative to the placement of the object on a support
surface, which is the dominant approach used for regrasp planning in the
literature).

We start by generating a set of unimanual grasps including the gripper positions
in $g_s$ and $g_t$ and other randomly sampled gripper positions.
We then combine these uni-manual grasps to also generate bimanual grasps.
Fig.~\ref{subfig:regraspsolution} represents the connectivity of these grasps as a
\textit{grasp graph}. Each node in the graph is a bimanual or uni-manual
grasp. A bimanual and a unimanual grasp is connected if the unimanual
grasp is one of the gripper poses in the bimanual grasp. Then, the planner can explore the graph to find a possible
path from  $g_{s}$ to  $g_{t}$, giving us the required sequence $G=(g_j)_{j=1}^{n}$.
This sequence consists of alternating bimanual and
unimanual grasps.  
Fig.~\ref{subfig:regraspsolution} highlights in red the shortest grasp sequence. There are other longer grasp sequences to connect $g_s$ and $g_t$ as well.

The grasp sequence acts as an abstract plan to guide the search in the lower
layers of the planner, and contracts the planning into a concrete and finite
group of grasp manifolds. 
In Alg.~1, the procedure $PlanRegrasp$ outlines this process. On lines 1-2, we
build the grasp graph and search it to generate the sequence of grasps $G$ as
outlined above. We then try to plan the motion from $q_s$ to $q_t$ through the
grasps $G$ (line 3).
If lower layers of our planner return with a failure to connect two grasps
$g_j$ and $g_{j+1}$ in $G$ (line 4), then we remove the link between these grasps in the
grasp graph (line 7), and perform the search again to generate a new sequence
of grasps (line 8). If the connection is successful, we return the re-grasp
motion to connect $q_s$ to $q_t$ (line 10).

%We generate these grasps by sampling from the grasp space a finite number of
%candidate grasps. We sample more if no solution from $g_{s}$ to $g_{t}$  is
%found.

\subsection{Sampling stable intersections of grasp manifolds}

%Given two subsequent
%        grasps from the sequence generated in the
%        layer above, we identify a single configuration $q$, at which the
%        transition from one grasp to the other can be performed
%        stably. 

        %This corresponds to identifying a sample
     %between the two grasp manifolds $M_{i}$, $M_{j}$, sample a set of feasibe configurations in their intersection $M_{i}\cap M_{j}$ to represent their connectivity.

A grasp path provides necessary but not sufficient conditions of the
connectivity of their corresponding grasp manifolds.  
To check this connectivity, given two subsequent grasps $g$ and $g'$, we
need to identify configurations at which both grasps are feasible and
stable against gravity.
Particularly in our task,
given the transition from a bimanual grasp to a unimanual grasp, the object may
not be stable against the gravity and slide within the remaining gripper.
Fig.~\ref{fig:subfig:circleinter} shows one such configuration in the bottom right corner which is not stable and a configuration, where the same transition is stable due
to a good choice of the configuration. Such configurations correspond to the
intersections of the two grasp manifolds that are stable against gravity.

%As shown in Fig.~\ref{fig:structure}, two neighboring grasp manifolds are connected if they
%intersect in $CS$ with a set of configurations at which one gripper can release
%off or regrasp the object. Meanwhile, every two neighboring intersections are
%connected by dual-arm or single-arm motions within grasp manifolds. Then, the
%bimanual regrasp planning is reduced to identifying the structure of the
%intersections among neighboring grasp manifolds (i.e., checking their connectivity
%) and then exploring the connectivity  of these intersections. 

%Sample-based planners can be applied to capture the structure of the
%intersection $M\left(g_{i}\right)\cap M\left(g_{i+1}\right)$  between two
%neighboring grasp manifolds.  In this paper, given two grasp manifolds, instead
%of exploring the complete structure of their intersection, our planner simply
In Alg.~1, the procedure $SampleIntersection$ samples $m$ such configurations.
To generate one such configuration, we first sample an object pose in the
reachable space of the robot (line 4). Then, we solve the inverse-kinematics
for the bimanual grasp at the sampled object pose, giving us a configuration
$q$ (line 5). We check (line 6) whether both grasps $g$ and $g'$ are stable
against gravity at $q$, using the same stability check described in
Sec.~\ref{sec:stabilitycheck}.  A stable configuration $q$ is returned as a
candidate point connection in the final solution path (line 7).

\subsection{Connectivity of sequence of manifold intersections}

%Check the connectivity of intersections among manifolds: closed-chain and single-arm motion planners are applied to check the connectivity of sampled configurations in intersections.
%The first layer of our planner identifies the configurations $Q$ that are
%stable against $F$, and then the third layer identifies intermediate
%configurations to perform the grasp changes in stable manner against gravity.

Given two configurations $q_s$ and $q_t$, and stable configurations sampled at
the intersections of a sequence of manifolds (i.e., the manifolds of the grasp
sequence $G$), we search for motion plans that connect $q_s$ to $q_t$ through
these manifolds.
%Each two subsequent configuration stays in a single manifold, and the path 
%segments connecting them 

In Alg.~1, the procedure $Connect$ implements this process as
depth-first-search.  Given a current configuration $q_s$ and a sequence of
grasps $G=(g_1,g_2,...,g_n)$ (where $g_1$ is the grasp in $q_s$), we sample the
intersection of the first two grasps in the sequence for stable configurations
(line 7). We then try to plan a motion from $q_s$ to a sampled configuration
$q$ (line 9).  Note that this is a motion plan within a single manifold (the
manifold of grasp $g_1$) and can be generated by existing closed-chain or
single-arm motion planners.  These paths, however, must also be stable against
gravity, for which constrained motion planners
\cite{berenson2011task,jaillet2013path} can be used.  If the motion plan is
successful, the trajectory is returned along with a recursive call to the
depth-first-search. Lines 1-6 handle the simple case where $q_s$ and $q_t$ are
already on the same manifold.

%Extending a path to an intermediate configuration is equivalent to connecting
%every two configurations in the same grasp manifold. It induces a
%pose-constrained manipulation planning problem that can be solved by existing
%sampled-based planners, such as CBiRRT \cite{berenson2011task}.
%
%To generate continuous regrasp motions, i.e., the connectivity of the
%intersections  among grasp manifolds, our planner uses depth-first search to
%find connected trajectory from $q_{s}$ to $q_{t}$ as shown from line 2 to line
%20 in function $ExploreRegraspGraph$. It recursively extends a path from $q_s$
%to an intermediate configuration in the subsequent intersection using a
%bimanual or unimanual transfer path until it reaches the target configuration
%$q_{t}$. 

 %\input{Regrasp/algorithm1}

%\input{Regrasp/Regrasp}
%\input{Feasibilitycheck/Feasibilitycheck}
%
%
%\input{Manipulation/Manipulation}

\section{Experiments and results}
\label{sec:Experiments} 

In this section, we present experiments to verify the performance of the proposed planners in terms of minimizing the number of regrasps and planning stable regrasps efficiently. The planners are applied to Baxter developed by Rethink Robotics in an OpenRAVE environment \cite{diankov2008openrave}. Baxter has two 7-DOF manipulators, each equipped with a  parallel jaw gripper.
We used a modified BiRRT planner \cite{kuffner2000rrt} as implemented in OpenRAVE as the motion planner to connect two configurations.
%The experiments start with validating the virtual joint based grasp feasibility checking method. Then, we present the performance of our multi-step planners in terms of minimizing the number of regrasps and planning efficiency. 

\begin{figure*}[tbp]
	\begin{center}
		\mbox{

			\subfigure{{
					\label{fig:subfig:measure_c}  
					\includegraphics[width=1.3 in, angle=-0]{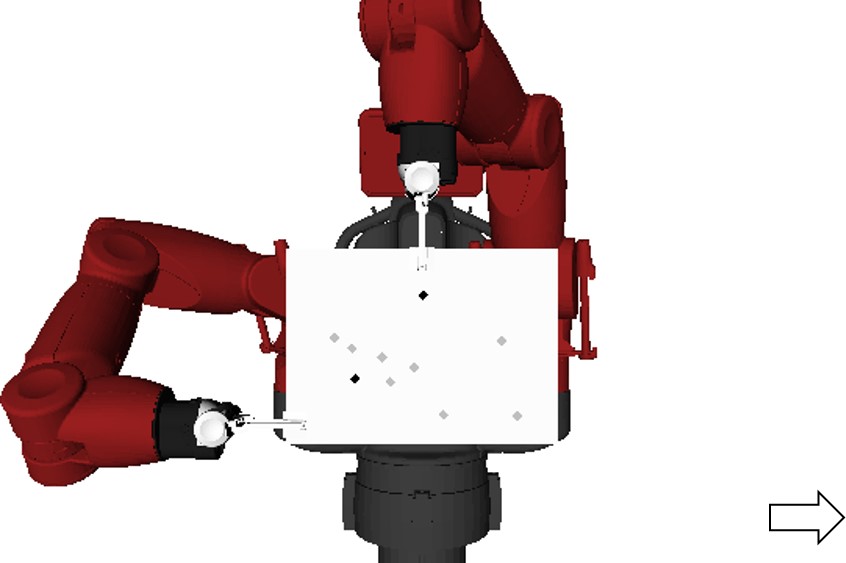}}}

			\subfigure{{
					\label{fig:subfig:measure_c}  
					\includegraphics[width=1.3 in, angle=-0]{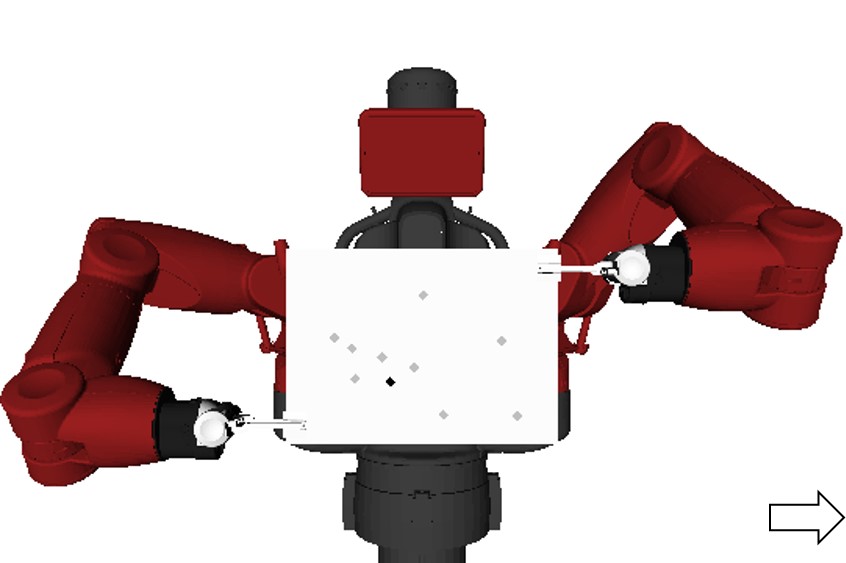}}}

			\subfigure{{
					\label{fig:subfig:A4-ran-force-1}  
					\includegraphics[width=1.3 in, angle=-0]{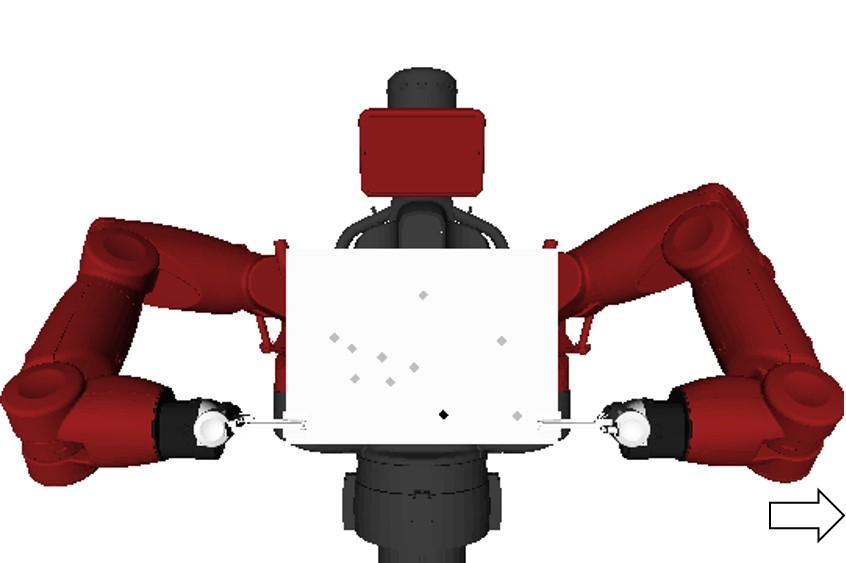}}}
			
			\subfigure{{
					\label{fig:subfig:measure_c}  
					\includegraphics[width=1.3 in, angle=-0]{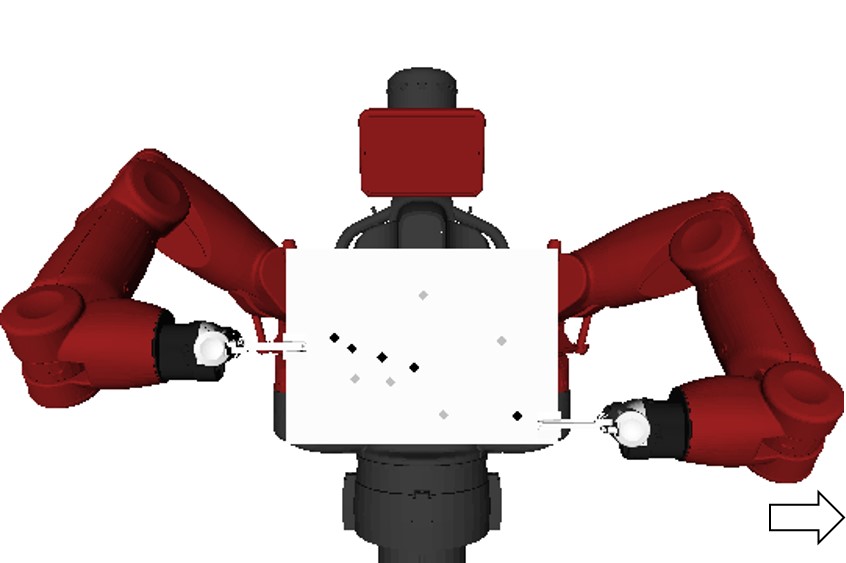}}}
			
			\subfigure{{
					\label{fig:subfig:measure_c}  
					\includegraphics[width=1.3 in, angle=-0]{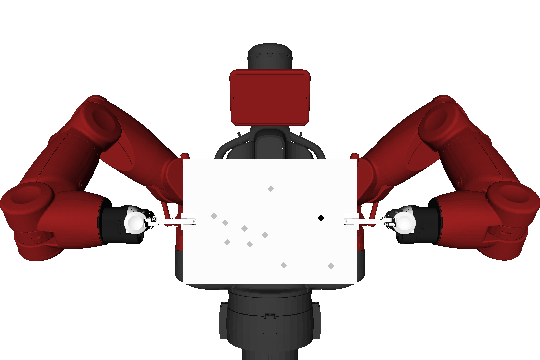}}}

		}

		\caption{A plan by the min-regrasp planner for a random-drilling task. The dark points indicate the drilling operations applied during the current grasp.}
		\label{fig:drilling-ran}
	\end{center}
\end{figure*}

The planners were tested on two types of forceful operations on a board, \textsl{drilling} and \textit{cutting}.
For all the drilling operations, we randomly changed the magnitude of the drilling forces from $10\,N$ to $15\,N$ and we assume the forces are normal to the surface of the board. For the cutting forces, we assume their magnitude varies between $30\,N$ to $60\,N$. 
These operations are instantiated into three categories of tasks, including:
\begin{itemize}
	\item \textit{random-drilling}: Each task contains 10 drilling operations randomly distributed on the surface of the board. An example is shown in Fig.~\ref{fig:drilling-ran};
	\item \textit{tick-drilling}: Each task contains 40 drilling operations along two random line segments meeting at a common point. An example is shown in Fig.~\ref{fig:real-tick};
	\item \textit{drilling\&cutting}: Each example contains four drilling operations and a cutting operation as shown in Fig.~\ref{fig:cutting}.
\end{itemize}

We generate 100 random tasks for each category above. 

In our experiments, we used a 
rigid foam board as the object. 
We also measured the force and torque limits
(as explained in Sec.~\ref{sec:stabilitycheck}) of the Baxter grippers on this object.
Along each axis shown in
Fig.~\ref{fig:forces}, we applied increasing amount of forces and torques to find
the point when the object started to slide between the parallel plates or when
the object rotated more than $5^{o}$ due to finger link deformation. We found the limits to be
$\left[{{P_x},P{}_y,{P_z},{R_x},{R_y},{R_z}} \right] = \left[13\,N, 40\,N, 13\,N, 0.3\,Nm, 0.05\,Nm, 0.1\,Nm\right]$.  Along the negative $P_z$ direction, the object rests
against the palm, therefore we used a large force limit ($100\,N$) in the
negative direction of $P_z$ when we solved Eq.~\ref{eq:virtual_cons}.

\begin{table}[]
	\centering
	\caption{Numbers of regrasps (with standard deviations) of three planners on three different tasks.}
	\label{table:minregrasp}
	\begin{tabular}{@{}cccc@{}}
		\toprule
		& Random-drilling & Tick-drilling & Drilling\&cutting \\ \midrule
		Random      & 17.6(0.9)       & 48.7(10.7)    & 5.8(2.1)         \\
		Greedy      & 7.8(1.9)        & 4.3(2.4)      & 3.1(0.8)          \\
		Min-regrasp & 5.2(0.9)        & 1.3(1.0)      & 2.0(0.0)          \\ \bottomrule
	\end{tabular}
\end{table}

\begin{figure*}[thp]
	\begin{center}
		\mbox{

			\subfigure{{
					\label{fig:subfig:measure_c}  
					\includegraphics[width=1.3 in, angle=-0]{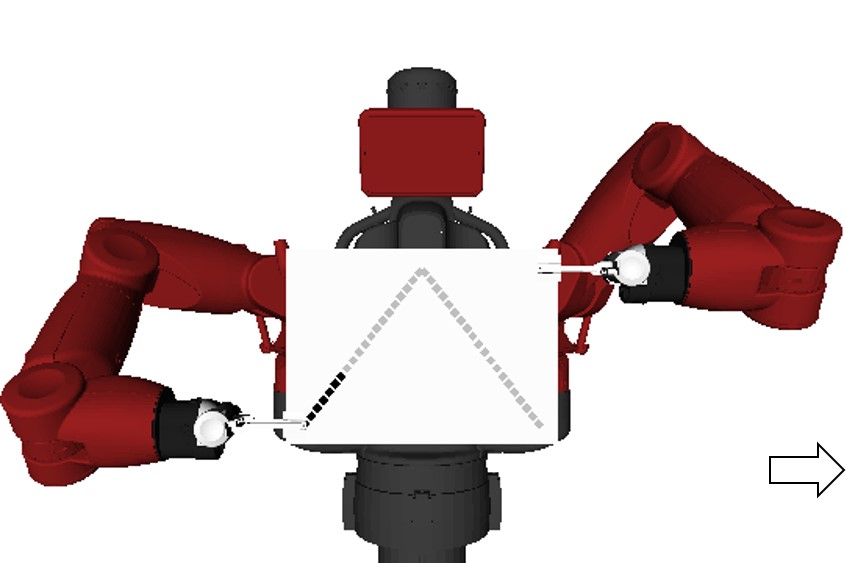}}}

			\subfigure{{
					\label{fig:subfig:measure_c}  
					\includegraphics[width=1.3 in, angle=-0]{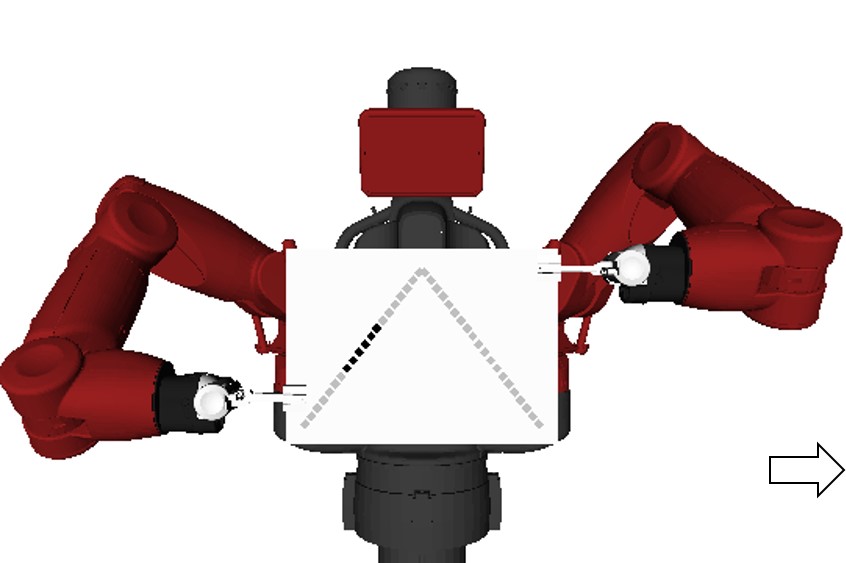}}}
			\subfigure{{
					\label{fig:subfig:measure_c}  
					\includegraphics[width=1.3 in, angle=-0]{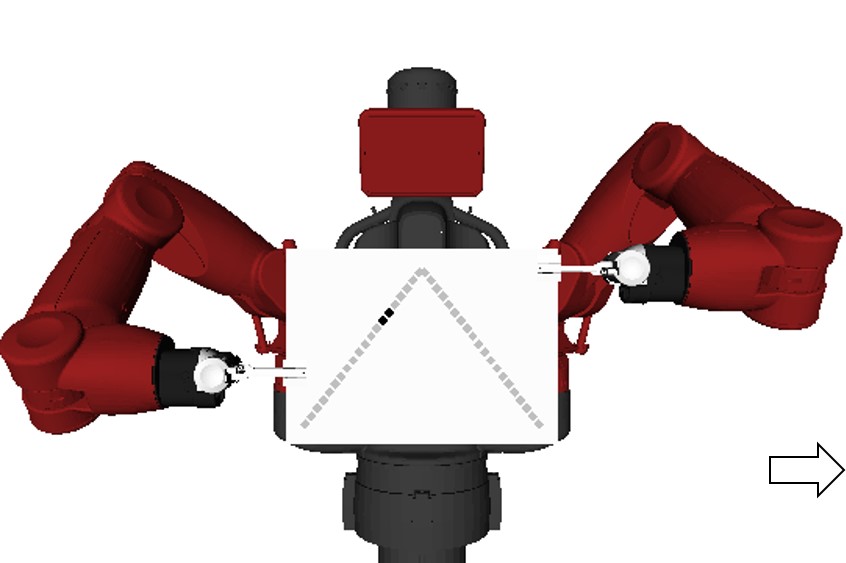}}}
		
			\subfigure{{
					\label{fig:subfig:measure_c}  
					\includegraphics[width=1.3 in, angle=-0]{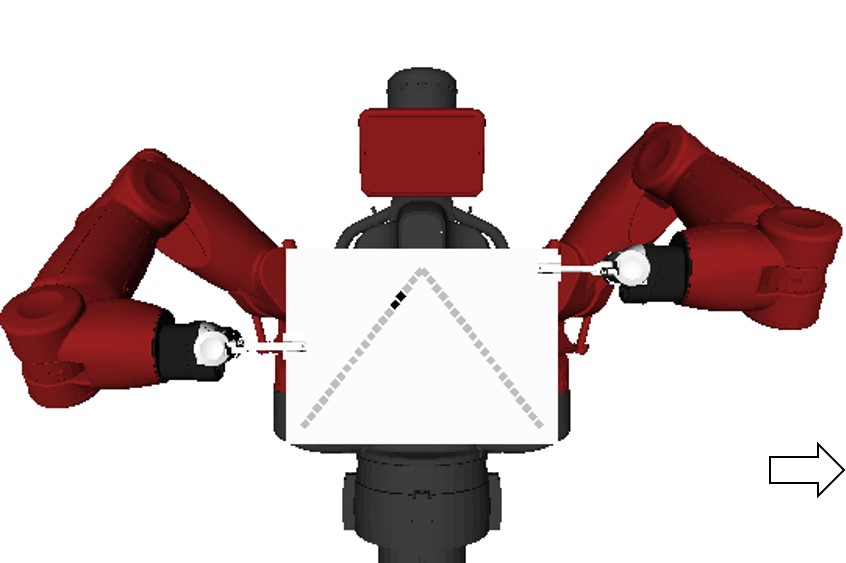}}}
			\subfigure{{
					\label{fig:subfig:measure_c}  
					\includegraphics[width=1.3 in, angle=-0]{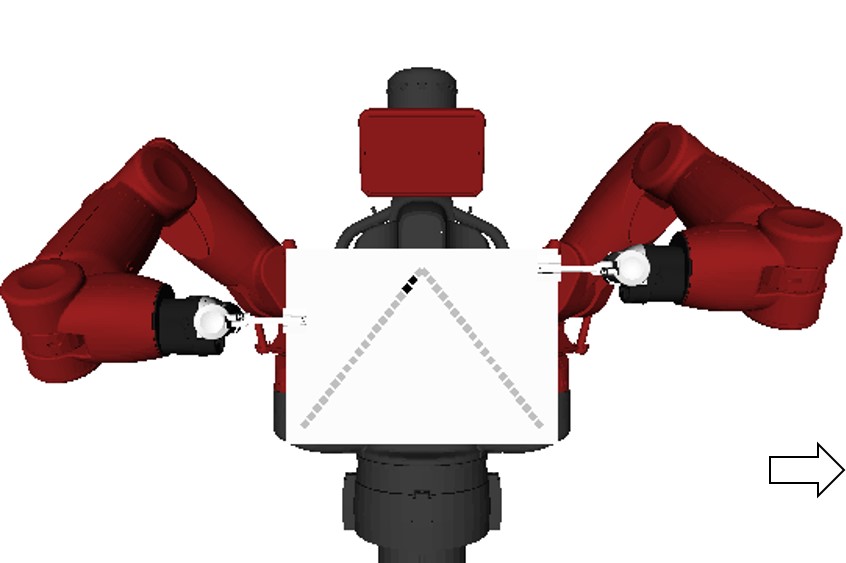}}}
			
		}

		\mbox{

			\subfigure{{
					\label{fig:subfig:measure_c}  
					\includegraphics[width=1.3 in, angle=-0]{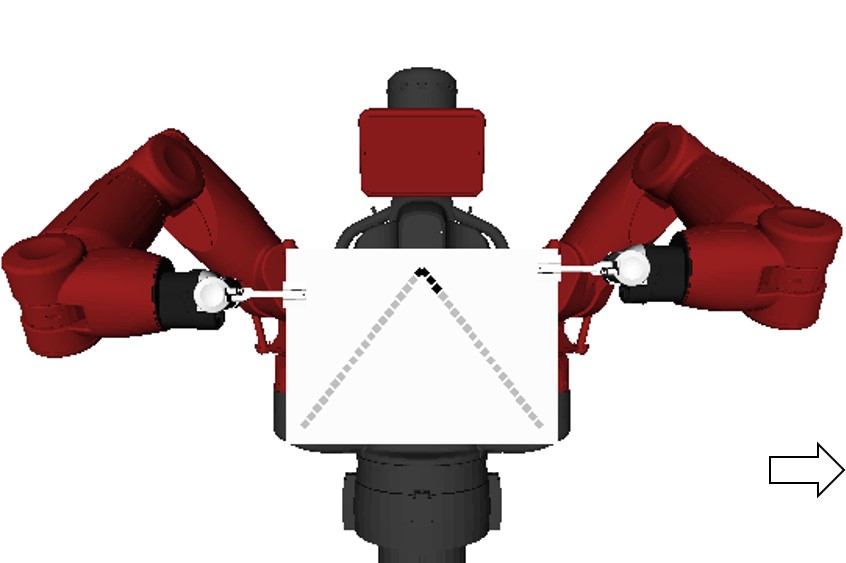}}}
			
			\subfigure{{
					\label{fig:subfig:measure_c}  
					\includegraphics[width=1.3 in, angle=-0]{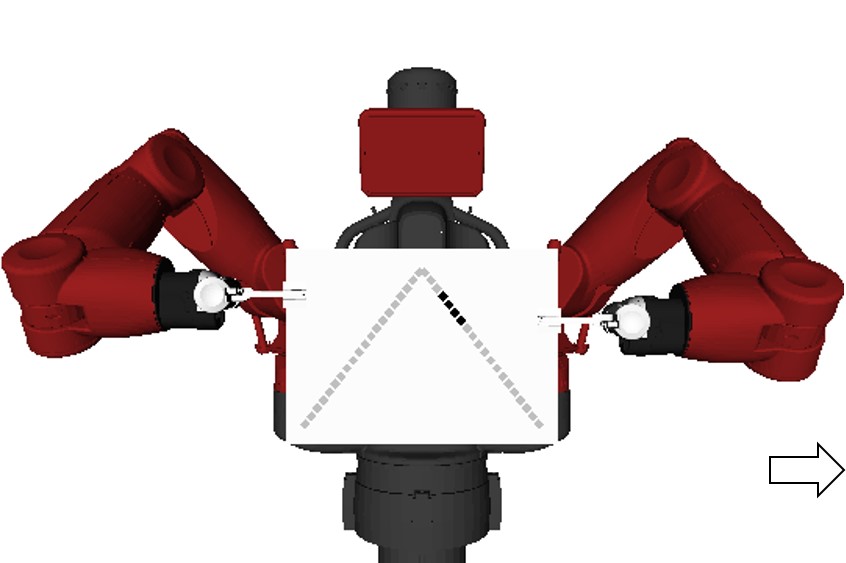}}}
			
			\subfigure{{
					\label{fig:subfig:measure_c}  
					\includegraphics[width=1.3 in, angle=-0]{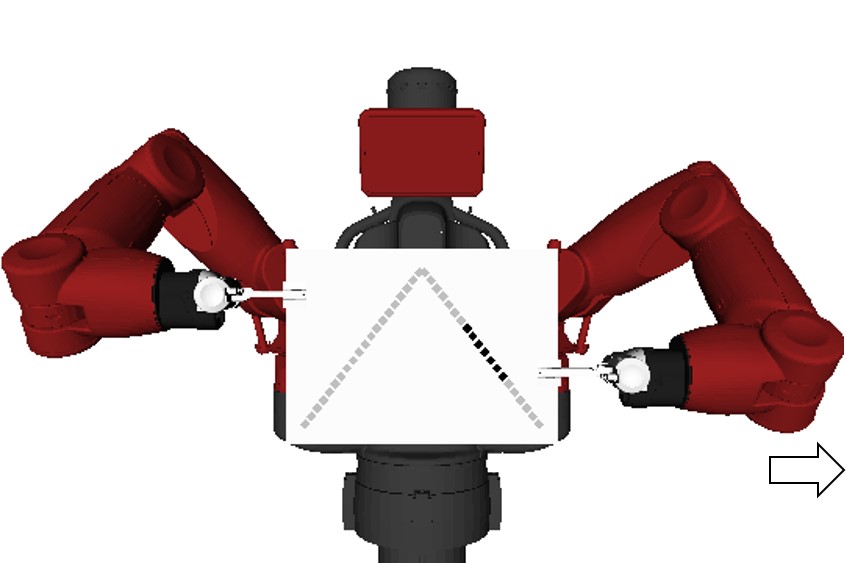}}}
			\subfigure{{
					\label{fig:subfig:measure_c}  
					\includegraphics[width=1.3 in, angle=-0]{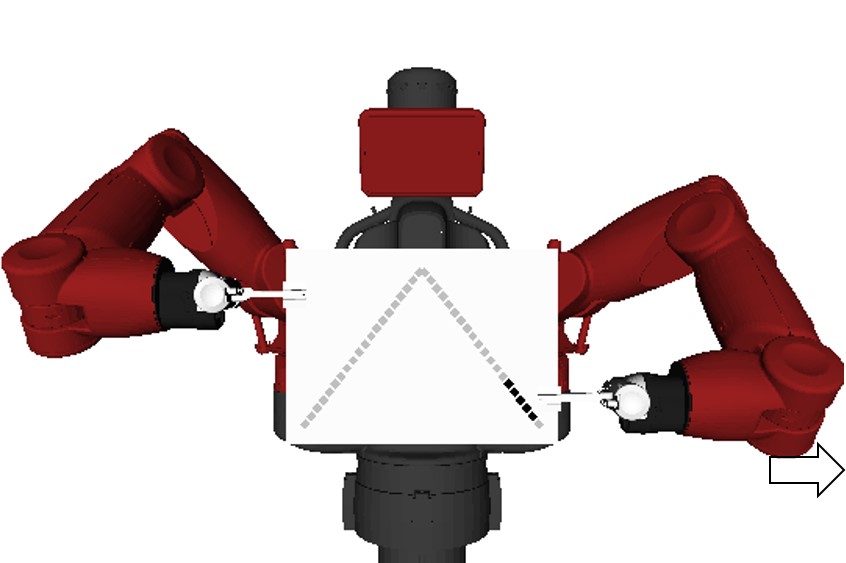}}}
			
			\subfigure{{
					\label{fig:subfig:measure_c}  
					\includegraphics[width=1.3 in, angle=-0]{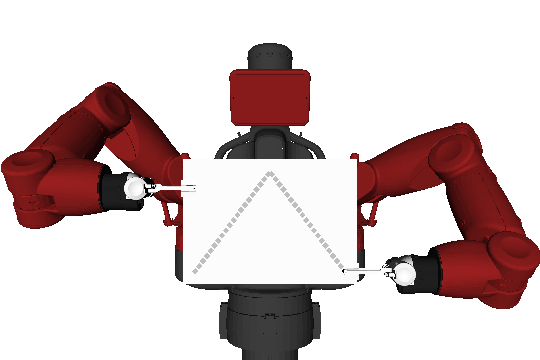}}}

		}

		\caption{A plan by the greedy planner for a tick-drilling task. The dark points indicate the drilling operations applied during the current grasp.}
		\label{fig:tick-greedy}
	\end{center}
\end{figure*}

\subsection{Minimizing the number of re-grasps}
\label{subsec:planningregrasp}
\begin{figure}[tbp]
	\begin{center}
		\mbox{

			\subfigure{{
					\label{fig:subfig:measure_c}  
					\includegraphics[width=1.6 in, angle=-0]{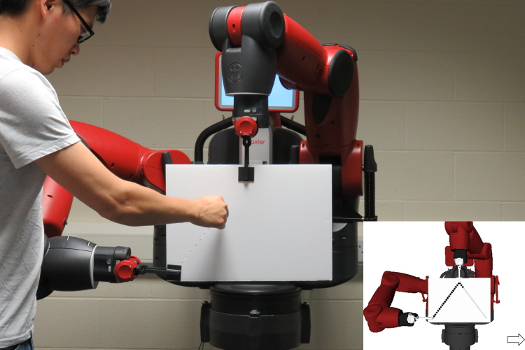}}}

			\subfigure{{
					\label{fig:subfig:measure_c}  
					\includegraphics[width=1.6 in, angle=-0]{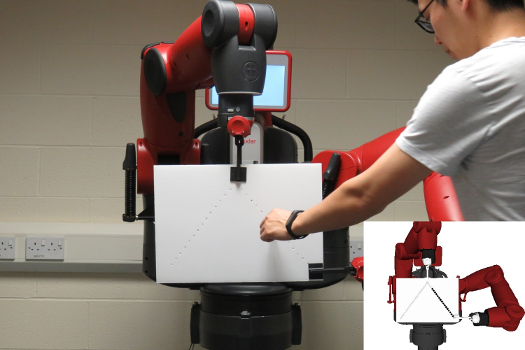}}}

		}

		\caption{A plan by the min-regrasp planner for a tick-drilling task.}
		\label{fig:real-tick}
	\end{center}
\end{figure}

First, we compared the performance of our planners, \textit{min-regrasp} and \textit{greedy}, with a random planner on the number of regrasps. The \textit{random planner} acts as a baseline approach. For the first
external force, the random planner samples a random configuration in the configuration space
until it finds a feasible one. For any subsequent force, it first checks
whether the configuration for the preceding force is still stable. If not, it falls back to random sampling. 

Table~\ref{table:minregrasp} shows the average results of the three planners on 100 random task instances. For the random-drilling tasks, the random planner generates almost one bimanual regrasp for every external force (maximum 18 regrasps for 10 external forces). The min-regrasp does dramatically minimize the number of regrasps (5 regrasps for 10 external forces, an example solution is shown in Fig.~\ref{fig:drilling-ran}).
The greedy planner also performs well in terms of reducing regrasps (8 regrasps).

Similarly, for the tick-drilling tasks, the random planner generates plans with a large number of regrasps (49 regrasps for 40 external forces of each tick-drilling task), while min-regrasp planner just needs 1.3 regrasps (an example solution is shown in Fig.~\ref{fig:real-tick}) and 2 regrasps for the drilling\&cutting tasks on average. The greedy planner shows a much better performance compared with the random planner, but still worse than the min-regrasp planner. For example, as shown in Fig.~\ref{fig:tick-greedy}, the greedy planner requires the grippers to climb along the edges of the board up and down frequently to follow the movements of the external forces, while the min-regrasp planner comes up with a plan of just two regrasps in Fig.~\ref{fig:real-tick}. We present a complete run of such a plan on the real robot in the attached video.

\begin{figure}[]
	\begin{center}
		\mbox{

			\subfigure{{
					\label{fig:subfig:measure_c}  
					\includegraphics[width=0.98 in, angle=-0]{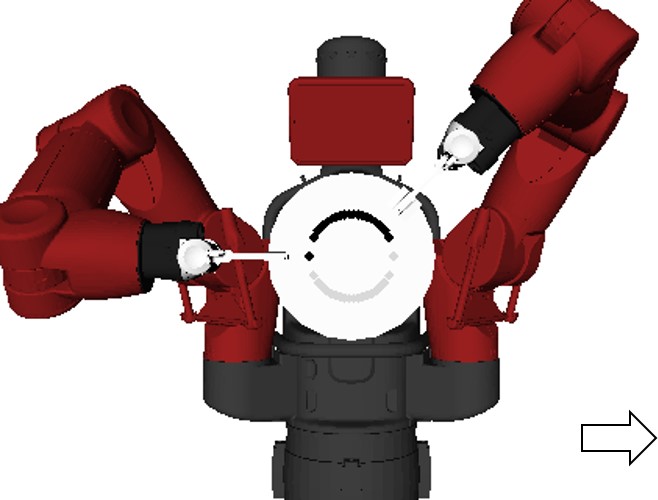}}}

			\subfigure{{
					\label{fig:subfig:measure_c}  
					\includegraphics[width=1.05 in, angle=-0]{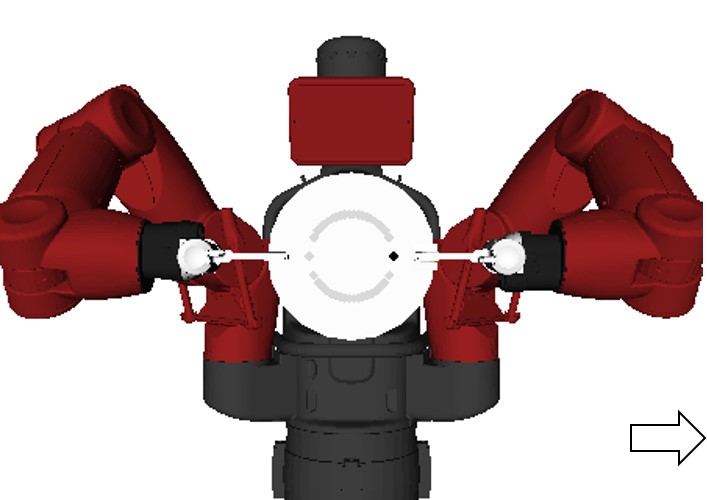}}}
			\subfigure{{
					\label{fig:subfig:measure_c}  
					\includegraphics[width=0.98 in, angle=-0]{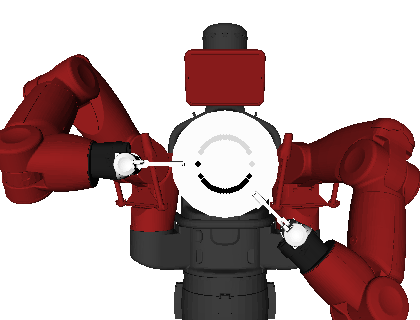}}}
			
		}

		\caption{A plan by the min-regrasp planner for drilling on a circular board. }
		\label{fig:circular}
	\end{center}
\end{figure}

We also counted the number of samples the random planner needed before it found
a feasible grasp. On average, the random planner needed \textbf{35.8 samples} for each
external force of the tasks above, showing that planning is necessary and random grasps have
little chance of being feasible. 
Our planners are not limited to grasping rectangular objects. To demonstrate
this, we tested the min-regrasp planner on a circular board with a sequence of
40 circular drilling operations. A plan with only two regrasps is shown in Fig.~\ref{fig:circular}.

\subsection{Planning performance}
\label{subsec:planningwithdrilling}

\begin{figure*}[tbp]
	\begin{center}
		\mbox{
			\subfigure[Start config.]{{
					\label{fig:subfig:m-i}  
					\includegraphics[width=0.9 in, angle=-0]{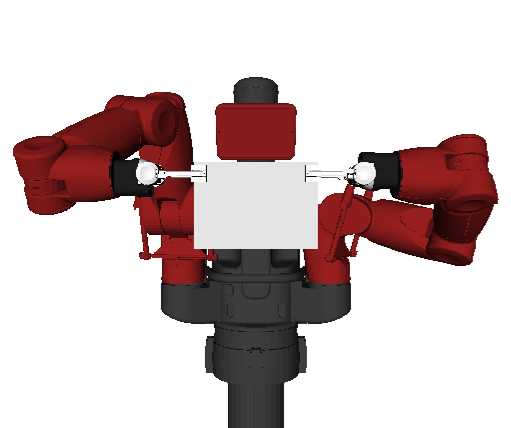}}}

			\subfigure[Intermed. config.]{{
					\label{fig:subfig:m-start}  
					\includegraphics[width=0.9 in, angle=-0]{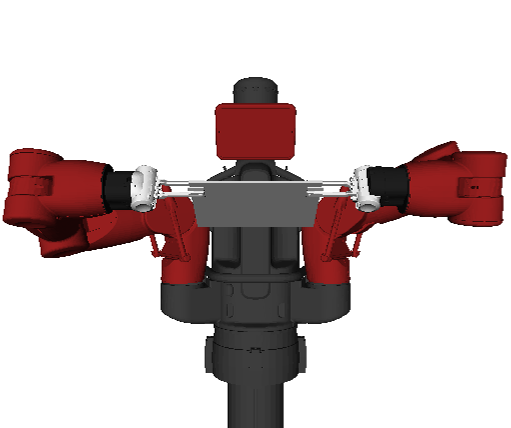}}}

			\subfigure[Release]{{
					\label{fig:subfig:m1}  
					\includegraphics[width=0.9 in, angle=-0]{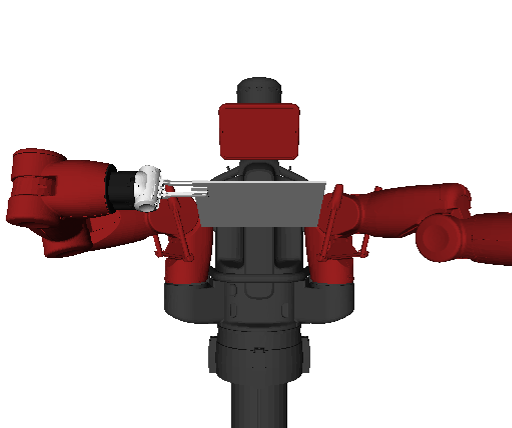}}}

			\subfigure[Regrasp]{{
					\label{fig:subfig:m2}  
					\includegraphics[width=0.9 in, angle=-0]{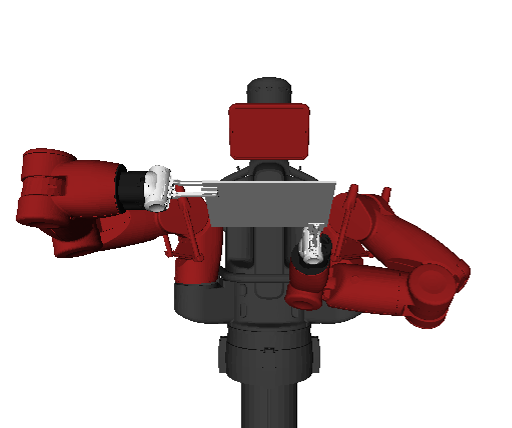}}}
			
			\subfigure[Release]{{
					\label{fig:subfig:m3}  
					\includegraphics[width=0.9 in, angle=-0]{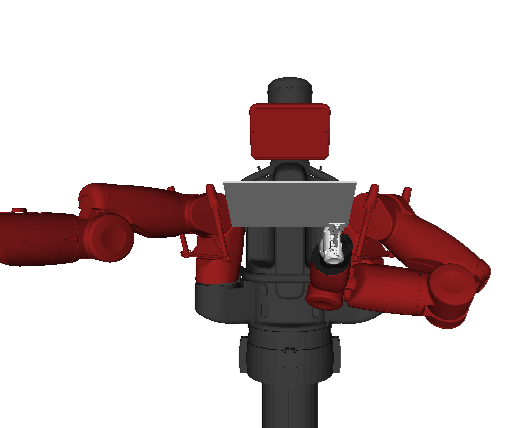}}}
			
			\subfigure[Regrasp]{{
					\label{fig:subfig:m-target}  
					\includegraphics[width=0.9 in, angle=-0]{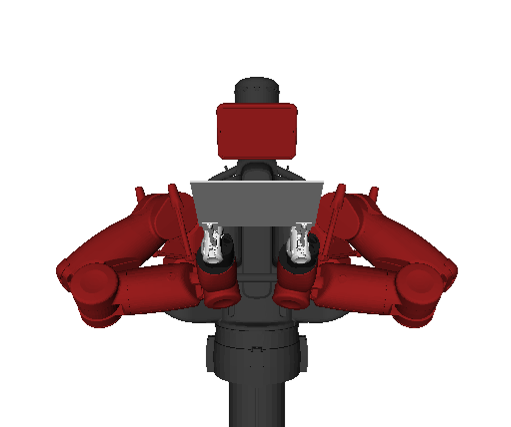}}}
			
			\subfigure[Target config.]{{
					\label{fig:subfig:m-t}  
					\includegraphics[width=0.9 in, angle=-0]{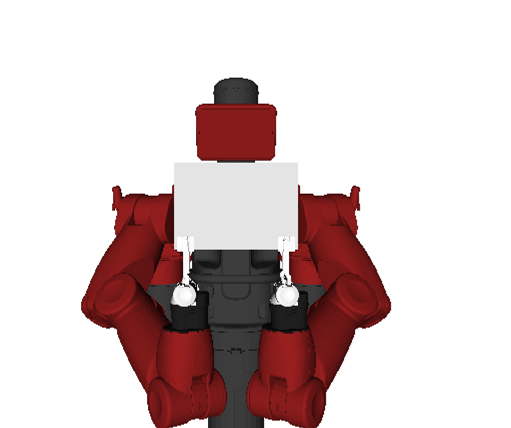}}}
										
		}
	
		\caption{Regrasping a heavy object.}
		\label{fig:m}
	\end{center}
\end{figure*}

We tested the performance of our planner on \textit{light} and \textit{heavy} objects respectively. 
We ran the planner on 100 randomly generated tasks for each category as
discussed above.
Table~\ref{table-minimizetime} shows the average planning time each layer of
the planner takes, including time for generating stable sequences (StabSeq for
short in Table~\ref{table-minimizetime}), time for generating and searching the grasp graph combined with sampling
intersections (SampInt, for short) and motion planning (Connect, for short).  
As the
table shows, most time is spent on motion planing, while the time for planning
stable configuration sequence and sampling intersection is negligible (The planner is set to generate a set of 20 feasible samples for each intersection).
Planning for the heavy object takes significantly long time because finding
stable regrasping configurations for this object is more difficult.

Fig.~\ref{fig:m} shows an example regrasp sequence  to regrasp a heavy object.  
For a light object, the robot can stably grasp and move the object using just a single gripper at most reachable configurations. 
%Fig~\ref{fig:light} shows one solution for re-grasping the light object.
Thus, mostly, the robot can directly release off and regrasp the object, without the need of reorienting it to intermediate configurations. 
However, for a heavy object, as discussed in Sec.~\ref{sec:intro}, the
object may slip down between gripper fingers if the robot directly releases one
gripper. That is, the robot needs to move it to intermediate configurations
at which one single gripper is  enough to keep the object stable.  In Fig.~\ref{fig:m}, the robot
first transfers the object to configurations in Fig.~\ref{fig:subfig:m-start}
and \ref{fig:subfig:m2} before releasing one gripper. After releasing, most
object weight will be resisted by the forces arising from gripper finger
bending as shown in Fig.~\ref{fig:subfig:m1} and \ref{fig:subfig:m3}, which are
much larger than the frictional forces between the object and finger surfaces.

\subsection{Real robot implementation}

\begin{figure}[tp]
	\begin{center}
		\mbox{
			\subfigure[Drill 1\&2 - Grasp 1]{{
					\label{fig:cutting:a}  
					\includegraphics[width=1.6in, angle=-0]{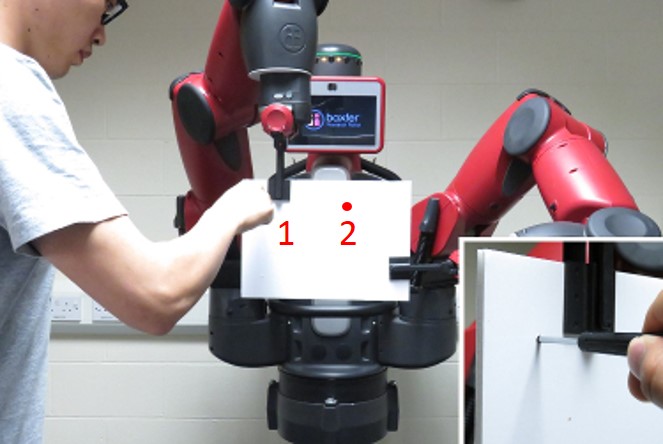}}}

			\subfigure[Drill 3 - Grasp 2 (after regrasp)]{{
					\label{fig:cutting:b}  
					\includegraphics[width=1.6 in, angle=-0]{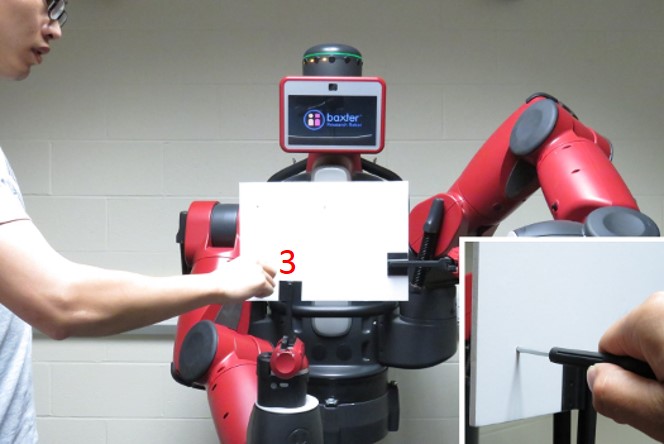}}}

		}

		\mbox{

			\subfigure[Drill 4 - Grasp 2]{{
					\label{fig:cutting:c}  
					\includegraphics[width=1.6 in, angle=-0]{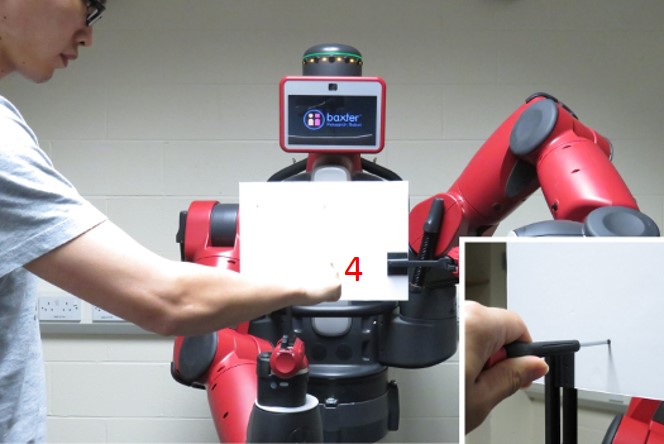}}}
									
			\subfigure[Cutting - Grasp 3 (after regrasp)]{{
					\label{fig:cutting:d}  
					\includegraphics[width=1.6 in, angle=-0]{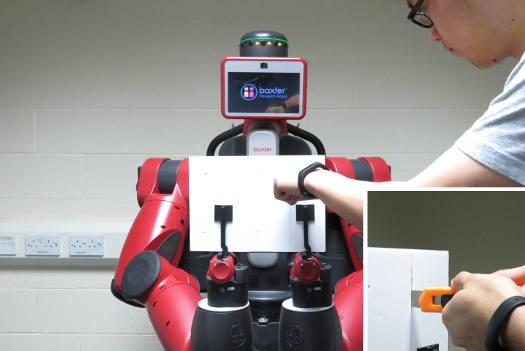}}}

		}

		\caption{Drilling\&cutting task. }
		\label{fig:cutting}
	\end{center}
\end{figure}

We ran our planner on a real Baxter robot for three tasks: cutting a circle, tick-drilling, and drilling\&cutting tasks.
The snapshots from these experiments are in Fig. \ref{fig:circlecut}, \ref{fig:real-tick} and \ref{fig:cutting}.
The attached video (\url{https://youtu.be/IHti307yGFY}) also presents these experiments .

% \usepackage{booktabs}
% \usepackage{multirow}
\begin{table*}[]
	\centering
	\caption{Planning time for both heavy and light objects. Times are in seconds. Standard deviations are in parantheses.}
	\label{table-minimizetime}
	\begin{tabular}{@{}ccclccccccccc@{}}
		\toprule
		\multirow{2}{*}{} & \multicolumn{4}{c}{random-drilling}                        &  & \multicolumn{3}{c}{tick-drilling}   &  & \multicolumn{3}{c}{drilling\&cutting} \\ \cmidrule(lr){2-5} \cmidrule(lr){7-9} \cmidrule(l){11-13} 
		& StabSeq     & \multicolumn{2}{c}{SampInt}   & Connect      &  & StabSeq   & SampInt   & Connect     &  & StabSeq   & SampInt    & Connect      \\ \cmidrule(r){1-5} \cmidrule(lr){7-9} \cmidrule(l){11-13} 
		heavy             & 11.2(2.5)   & \multicolumn{2}{c}{50.1(4.7)} & 440.0(62.3)  &  & 11.7(0.8) & 12.8(1.0) & 114.4(17.3) &  & 2.2(0.2)  & 20.6(1.2)  & 139.1(25.0)  \\
		light             & 10.9(2.8)   & \multicolumn{2}{c}{17.8(1.9)} & 155.5(11.4)  &  & 11.9(0.8) & 5.0(0.3)  & 39.8(8.3)   &  & 1.9(0.4)  & 6.6(0.7)   & 71(14.1)     \\ \bottomrule
	\end{tabular}
\end{table*}

\section{Conclusion and Future Work}

We believe the planning system presented here can be a key component in a
human-robot collaboration framework.  In future work, we aim to include an
increasing amount of human comfort factors (e.g. the human kinematics) in
planning the collaboration between the human and the robot.

\bibliographystyle{IEEEtran}
\bibliography{References/references}

\end{document}